The paper may be considered for *(indicate your choice by putting √ in the appropriate box)*

| | |
|---|---|
| *1. Oral Presentation* | √ |
| *2. Poster Session* | |

# Permeability Analysis based on information granulation theory

M.Sharifzadeh, H.Owladeghaffari, K.Shahriar, E.Bakhtavar,
*Dept. of Mining and Metallurgical Engineering, Amirkabir University of Technology, Tehran, Iran*

**Keywords: Information granulation theory, SOM, NFIS, RST, permeability**

ABSTRACT: This paper describes application of information granulation theory, on the analysis of "lugeon data". In this manner, using a combining of Self Organizing Map (SOM) and Neuro-Fuzzy Inference System (NFIS), crisp and fuzzy granules are obtained. Balancing of crisp granules and sub- fuzzy granules, within non fuzzy information (initial granulation), is rendered in open-close iteration. Using two criteria, "simplicity of rules "and "suitable adaptive threshold error level", stability of algorithm is guaranteed. In other part of paper, rough set theory (RST), to approximate analysis, has been employed .Validation of the proposed methods, on the large data set of in-situ permeability in rock masses, in the Shivashan dam, Iran, has been highlighted. By the implementation of the proposed algorithm on the lugeon data set, was proved the suggested method, relating the approximate analysis on the permeability, could be applied.

## 1 Introduction

During the dam structures design, one of the most significant issues is the estimation of permeability variations in different levels of the dam site. However, prediction of permeability, using obtained data, from in-situ tests is a big challenge. Relating to the determination of potential water flow paths within the rock mass, underlying a potential dam structure is especially important and this has an extensive impact on the planning of grouting procedures (Houlsby 1990). Several different methods for assessing the permeability variations in the rock mass have been reviewed in the literature: Nakaya et al. (1997) and Shahriar & Owladeghaffari (2007).

Due to association of uncertainty and vagueness with the monitored data set, particularly, resulted from the in-situ tests (such lugeon test), accounting relevant approaches such probability, fuzzy set and rough set theories to knowledge acquisition, extraction of rules and prediction of unknown cases, more than the past have been distinguished. The Information Granulation (IG) theory covers the mentioned approaches in two formats: crisp (no-fuzzy) IG and fuzzy IG (Zadeh, 1997). There are two main reasons why we propose IG theory to tackle with uncertainty in the monitored geomechanics data. The first one is human instinct. As human being, a granular view of the world has been developed. In this study, using two Computational Intelligence (CI) theories namely neural networks, and fuzzy inference system, based on IG theory, an algorithm to analysis permeability data was presented and applied to the Shivashan dam site located in north western of Iran. Other part of study investigates application of Rough Set Theory (RST), as a new approximate analysis, on these data set and comparison of results with former algorithm. In first model, self-organizing feature map and Neuro-Fuzzy Inference System is utilized to construct IGs. To determine suitable granulation level, the two criteria, "simplicity of rules" and "adaptive threshold error level", are supposed.

## 2  Construction of information granules

Information granules are collections of entities that are arranged due to their similarity, functional adjacency, or indiscernibility relation. The process of forming information granules is referred to as IG. There are many approaches to construction of IG, for example SOM, Fuzzy C-Means (FCM), and RST. The granulation level depends on the requirements of the project. The smaller IGs come from more detailed processing. On the other hand, because of complex innate feature of information in real world and to deal with vagueness, adopting of fuzzy and rough analysis or the combination form of them is necessary. In this study, the main aim is to develop a hierarchical extraction of IGs using three main steps:

1-Random selection of initial crisp granules: this step can be set as "Close World" Assumption .But in many applications, the assumption of complete information is not feasible (CWA), and only cannot be used.  In such cases, an Open World Assumption (OWA), where information not known by an agent is assumed to be unknown, is often accepted (Dohert et al, 2007).

2- Fuzzy granulation of initial granules: sub fuzzy granules inside precise granules and extraction of if-then rules.

3- The close-open iteration: this process is a guideline to balancing of crisp and sub fuzzy granules by some random selection of initial granules or other optimal structures and increment of supporting rules, gradually. This paper employed two main approaches on constructing of IGs: self organizing feature map as initial granulation, and NFIS as secondary granulation. Other process, in this manner, is applying of RST.

### 2.1  Self Organizing Map-neural network (SOM)

Kohonen self-organizing networks (Kohonen feature maps or topology-preserving maps) are competition-based network paradigm for data clustering. The learning procedure of Kohonen feature maps is similar to the competitive learning networks. The main idea behind competitive learning is simple; the winner takes all. The competitive transfer function returns neural outputs of 0 for all neurons except for the winner which receives the highest net input with output 1.

SOM changes all weight vectors of neurons in the near vicinity of the winner neuron towards the input vector. Due to this property SOM, are used to reduce the dimensionality of complex data (data clustering).  Competitive layers will automatically learn to classify input vectors, the classes that the competitive layer finds are depend only on the distances between input vectors (Kohonen, 1990).

### 2.2  Neuro-Fuzzy Inference System (NFIS)

There are different solutions of fuzzy inference systems. Two well-known fuzzy modelling methods are the Tsukamoto fuzzy model and Takagi– Sugeno–Kang (TSK) model. In this study, only the TSK model has been considered. A fuzzy rule in this model has following form:

$$if\ x_i\ is\ A_1\ and\ x_2\ is\ A_2\ \ and\ \ x_n\ is\ A_n\ \ then\ y = f(x) \tag{1}$$

where $f(x)$ is crisp function in the consequent. The function $y=f(x)$ is a polynomial in the input variables $x_1$, $x_2$, …,$x_n$.We will apply here the linear form of this function. For $M$ fuzzy rules of the equation 1, we have $M$ such membership functions $\mu_1, \mu_2,…, \mu_M$ We assume that each decision part is ensued by the consequent of the linear form as the equation 2:

$$= P_{i0} + \sum_{j=1}^{n} P_{ij}\, x_j \quad i = 1, 2,...,M \ \ and \ \ j = 1, 2,...,n \tag{2}$$

The algebraic product aggregation of the input variables, at the existence of $M$ rules, the Neuro–fuzzy TSK system output signal $y(x)$ upon excitation by the vector $x$ is described by the equation 1.

The adjusted parameters of the system are the nonlinear parameters $\left(c_j^{(k)}, \sigma_j^{(k)}, b_j^{(k)}\right)$ for $j = 1, 2,..., n$ and $k = 1, 2, ..., M$ of the fuzzifier functions and the linear parameters (weights $P_{kj}$) of TSK functions. In contrary to the Mamdani fuzzy inference system, the TSK model generates a crisp output value instead of a fuzzy one. The defuzzifier is not necessary.
The TSK fuzzy inference systems described by equation 3 can be easily implanted in the form of a so called Neuro-fuzzy network structure.

$$y(x) = \frac{1}{\sum_{r=1}^{M}\left[\prod_{j=1}^{n}\mu_r\left(x_j\right)\right]} \times \sum_{k=1}^{M}\left(\left[\prod_{j=1}^{n}\mu_k\left(x_j\right)\right]\left(p_{k0}+\sum_{j=1}^{n}p_{kj}x_j\right)\right) \tag{3}$$

Figure 1 presents the *5-layer* structure of a Neuro-fuzzy network, realizing the TSK model of the fuzzy system. It is assumed that the functions $y_i$, $y_i = f_i(x)$ are linear of the form (as equation 4)

$$f_i(x) = p_{i0} + \sum_{j=1}^{n}p_{ij}x_j \tag{4}$$

The parameters of the networks are the variables of the membership functions $\left(c_j^{(k)}, \sigma_j^{(k)}, b_j^{(k)}\right)$ *for j = 1, 2,..., n and k = 1, 2, ..., M* and the coefficients (linear weights) $p_{ij}$ *for i =1,2,...,M and j =0,1,2,...,n* of the linear Takagi–Sugeno functions. The network in figure 1 has a multilayer form, in which any inputs( *x,y*), as condition attributes, has two MFs.  Details of the procedure can be found in(Jang et al, 1997).

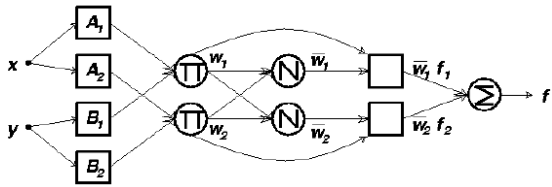

Figure 1. A typical ANFIS (TSK) with two inputs and two MF for any input (Jang et al, 1997)

### 2.3  Rough Set Theory (RST)

The rough set theory introduced by Pawlak (Pawlak, 1991) has often proved to be an excellent mathematical tool for the analysis of a vague description of object.  The adjective vague referring to the quality of information means inconsistency, or ambiguity which follows from information granulation.
An information system is a pair *S*=< *U, A* >, where *U* is a nonempty finite set called the universe and *A* is a nonempty finite set of attributes.  An attribute a can be regarded as a function from the domain *U* to some value set $V_a$. An information system can be represented as an attribute-value table, in which rows are labeled by objects of the universe and columns by attributes.  With every subset of attributes $B \subseteq A$, one can easily associate an equivalence relation $I_B$ on U:

$$I_B = \{(x, y) \in U : for\ every\ a \in B, a(x) = a(y)\} \tag{5}$$

Then, $I_B = \bigcap_{a \in B} I_a$.

If $X \subseteq U$ ,the sets $\{x \in U : [x]_B \subseteq X\}$ and $\{x \in U : [x]_B \cap X \neq \varphi\}$, where $[x]_B$ denotes the equivalence class of the object $x \in U$ relative to $I_B$, are called the B-lower and the B-upper approximation of *X* in *S* and denoted by $\underline{BX}$ and $\overline{BX}$, respectively.  Consider $U = \{x_1, x_2, ..., x_n\}$ and $A = \{a_1, a_2, ..., a_n\}$ in the information system $S = \prec U, A \succ$. By the discernibility matrix M(S) of S is meant an n*n matrix such that:

$$c_{ij} = \left\{ a \in A : a(x_i) \neq a(x_j) \right\} \tag{6}$$

A discernibilty function $f_s$ is a function of m Boolean variables $a_1 ... a_m$ corresponding to attribute $a_1 ... a_m$, respectively, and defined as follows:

$$f_s(a_1, ..., a_m) = \wedge \{ \vee (c_{ij}) : i, j \leq n, j \prec i, c_{ij} \neq \varphi \} \tag{7}$$

where $\vee (c_{ij})$ is the disjunction of all variables with $a \in c_{ij}$. Using such discriminant matrix the appropriate rules are elicited (Pal&Mitra, 2004). In this study we have developed dependency rule generation –RST- in MatLab7, and on this added toolbox other appropriate algorithms have been prepared.

## 3   The proposed procedure based on balancing of granules

In this section, utilizing Information Granulation (IG) theory, a new procedure is described, to demonstrate permeability varirtions in the unlike sections. Figure 2 illustrates the basic idea of the proposed methodology. The detailed procedure is respectively described as following:
Step (1): dividing the monitored data into groups of training and testing data
Step (2): first granulation (crisp) by SOM
         Step (2-1): selecting the level of granularity randomly.
         Step (2-2): construction of the granules (crisp).
Step (3): second granulation (fuzzy IG) by NFIS
         Step (3-1): crisp granules as a new data.
         Step (3-2): selecting the level of granularity. (Error level-number of rules)
         Step (3-3): checking the suitability. (Close-open iteration)
         Step (3-4): construction of fuzzy granules.
Step (4): extraction of knowledge rules
Obviously, the granularity level is controlled by Error Level (EL), number of rules and number of neurons in crisp granules. So, the latter criteria are based on "simplicity reasons for nature events" in human treatment, so that, the world's cognition may be granulated, organized and caused with minimum and simplest rules. By considering the subtractive clustering method, to create fuzzy granules, the role of influence radius in partition of data set, and by suitable selection of this range, algorithm is started.

## 4   Dam permeability

Shivashan hydroelectric earth dam is located 45km north of Sardasht city in north western of Iran.
In order to obtain engineering geological information, boreholes were drilled at different points of Shivashan dam's area. Water Pressure Test (WPT) has used for determination of this area's permeability. WPT is an effective method for widely determination of rock mass permeability totally, 20 boreholes have been drilled and consequently about 789 data set were resulted.
To evaluate the permeability due to the lugeon values and the proposed method, two different cases were considered: first case is on the local coordinates of dam site (position of boreholes: *x,y,z*) to depict 3D isolugeon diagrams ;while in other step aim is to detect relation between available measured data from boreholes. In latter case, the input parameters were selected as follows: Z (elevation of any section), ΔL (length of tested section), RQD, TWR (type of weathering rock).
In first analysis, without using proposed algorithm, by direct solution of ANFIS and using 3MFs for any inputs, prediction of permeability in different levels (Z=1160,1180,1190 and 1200) was accomplished (figure 3a). Same process based on the deterministic extracted rules from RST, and transferring of attributes by SOM in 5 symbolic levels: very high, high, medium, low and very low was rendered in different levels (Figure 3b). In RST analysis, the symbolic values of 1(very low)… 5(very high), 6(no-deterministic), were attributed to the lugeon values. In Contrary one may interprets the variations in z= {z*} is the superposition of the sub levels results, has been emerged by NFIS, approximately. In figure 4 variations of RQD by NFIS and using five *MFs* has been portrayed.

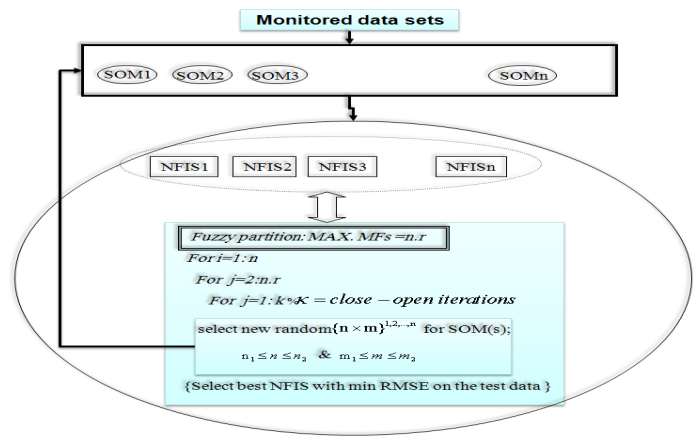

Figure 2. The procedure of crisp-sub fuzzy granulation

Second step was based on the proposed method. The effective parameters in the proposed algorithm are: *n=1* (number of random initial SOM&NFIS); *k =3* (maximum closed-open iterations for each rule), *n.r*=4 (maximum rules) and ξ≈23 (error level). After two open-close iterations, the conditions of algorithm were satisfied; so that by 10*15 SOM crisp granules were obtained. (*n=10, m=15*; Matrix of neurons (*n * m*) determines the size of 2D SOM).

Indeed, SOM can be supposed as a pre-processing step (comparison between figures 5a-b).

The obtained results in step 2 were stored in the aggregated data box, so that according to the mentioned data box the estimation of optimum point of balancing between crisp and fuzzy granules can be done.

Best structure of explored NFIS, presents the dominant rules as follow:

$1. If\ (x_1 \in in1m\ f1)\ \&\ (x_2 \in in2m\ f1)\ \&\ (x_3 \in in3m\ f1)\ \&\ (x_4 \in in4m\ f1) then\ output1 = f1$

$2. If\ (x_1 \in in1m\ f2)\ \&\ (x_2 \in in2m\ f2)\ \&\ (x_3 \in in3m\ f2)\ \&\ (x_4 \in in4m\ f2) then\ output\ 2 = f2$

$3. If\ (x_1 \in in1m\ f3)\ \&\ (x_2 \in in2m\ f3)\ \&\ (x_3 \in in3m\ f3)\ \&\ (x_4 \in in4m\ f3) then\ output3 = f3$

$4. If\ (x_1 \in in1m\ f4)\ \&\ (x_2 \in in2m\ f4)\ \&\ (x_3 \in in3m\ f4)\ \&\ (x_4 \in in4m\ f4) then\ output\ 4 = f4$

Where input parameters (xi) belong to the Gaussian format of membership functions (figures 6.a, b, and c, d). As well as, we can write the linear formats of decision attribute (sub-fuzzy) based on conditional parameters, which are as following:

$$f_1 = p_1 x_1 + \ldots + p_4 x_4 + r_1$$
$$p_1 = 0.0186 \; ; \; p_2 = 14.2825 ; p_3 = \text{-}0.0657 \; ; p_4 = 11.6620 \; ; r_1 = \text{-}38.4240$$
$$f_2 = q_1 x_1 + \ldots + q_4 x_4 + r_2$$
$$q_1 = 0.0928 \; ; \; q_2 = 7.1981 ; q_3 = 2.9592 \; ; q_4 = 2.4851 \; ; r_2 = \text{-}117.691$$
$$f_3 = s_1 x_1 + \ldots + s_4 x_4 + r_3$$
$$s_1 = \text{-}0.0274 \; ; \; s_2 = \text{-}8.0836 ; s_3 = 8 \; 0.0694 \; ; s_4 = 11.4711 \; ; r_3 = 42.4835$$
$$f_4 = t_1 x_1 + \ldots + t_4 x_4 + r_4$$
$$t_1 = \text{-}0.2195 \; ; \; t_2 = 98.0854 ; t_3 = \text{-}0.0816 \; ; t_4 = 20.0027 \; ; r_4 = 138.4553$$

The rules state the relationship between inputs and output in 3D. For instance figure 7 demonstrates general and possible variation of lugeon with RQD and Z: with increasing RQD, lugeon is decreased while high elevations are coincided by high lugeon. Such features may be evaluated, with more details, using figure5b, where the scatter training data have transferred in to 150 data. For example, three major patterns in lugeo-RQD or lugeon-T.W.R confirm three main unlike treatments of the rock mass, induced from the different patterns of joints and filling materials.

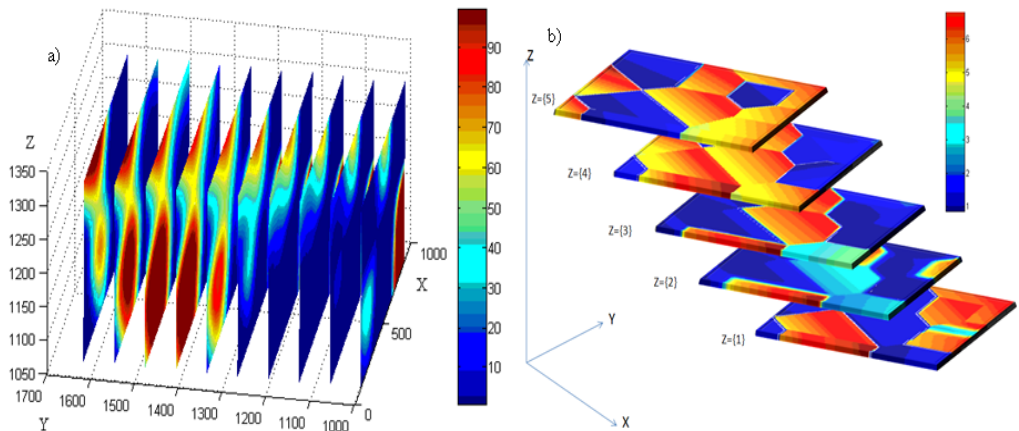

Figure 3.a) Isolugeon graphs by NFIS, b) RST Performance in symbolic levels and five scaling of attributes. Number 6(more than 5) characterizes ambiguity and unknown cases

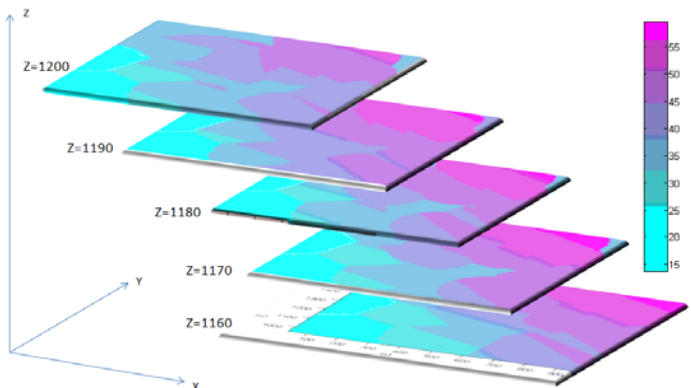

Figure4. Iso-surfaces of RQD by NFIS

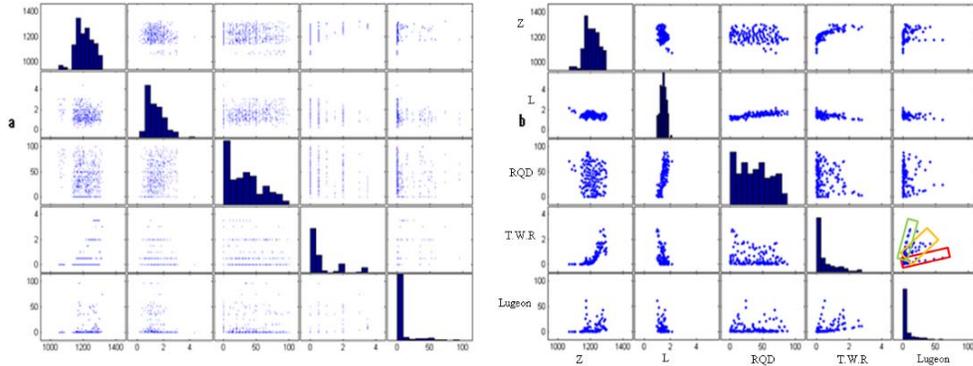

Figure5. a) Real data set in matrix plot form (as training data set); b) matrix plot of crisp granules by 10*15 SOM after 500 training

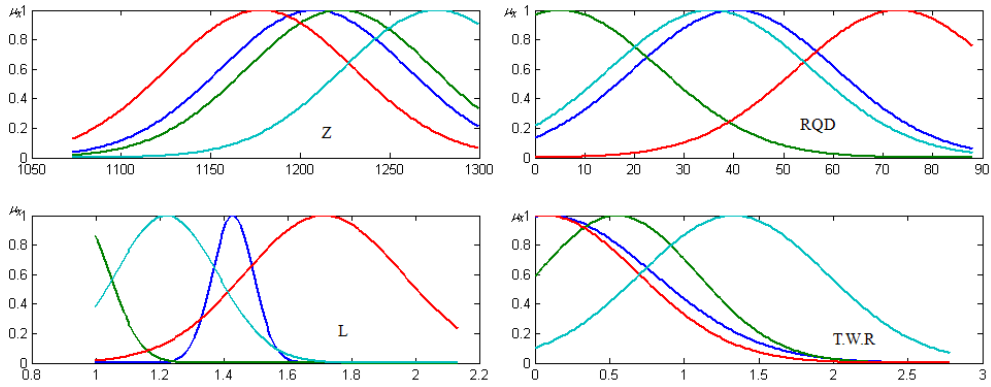

Figure 6.a, b, c, d. Fuzzy granulation of conditional attributes (inputs), - vertical axis shows degree of membership function

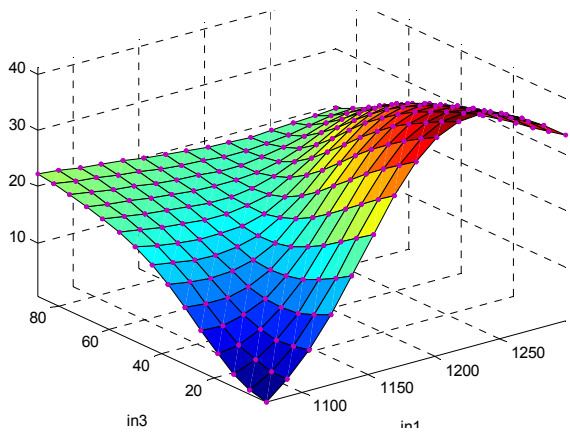

Figure 7. The 3D relation between *Z (in1), RQD (in3), and lugeon*

## 5 Conclusion

Uncertainty and vague information has undeniable role in geomechanic's analysis. Indeed, developing of new approaches in data engineering and computational intelligence, as well as, natural computing approaches, it is necessary to consider such approaches to better understand of natural events in rock mass. Under this idea and to find out the best information granules (clusters) which have intricate structures, close-open worlds (cycle) procedure to balancing of successive granules was proposed. By the implementation of the proposed algorithm on the lugeon data set of Shivashan dam, was proved the suggested method relating the approximate analysis on the permeability could be applied. From the mentioned analysis the following results can be deduced:

1- Detection of the permeability variations in successive level using NFIS and RST: the permeability in left bank is lower than other side.
2- Elicitation of best simple rules between effective parameters
3- A pre-processing on the scatter lugeon data using best SOM-in balancing with NFIS

Applications of new data engineering methods in geomechanics and combining with hard computing, under new algorithms, are future works of authors.